\documentclass{article}

\usepackage{arxiv}

\usepackage[utf8]{inputenc} 
\usepackage[T1]{fontenc}    
\usepackage{hyperref}       
\usepackage{url}            
\usepackage{booktabs}       
\usepackage{amsfonts}       
\usepackage{nicefrac}       
\usepackage{microtype}      
\usepackage{lipsum}

\usepackage{bm}
\usepackage{amsmath}
\usepackage{amssymb}
\usepackage{amsfonts}
\usepackage{graphicx}
\def\vec#1{\mathbf{#1}}
\usepackage{algorithm}
\usepackage{algorithmic}
\usepackage{comment}

\title{Few-shot Learning for Spatial Regression}

\author{
  Tomoharu Iwata\\
  NTT Communication Science Laboratories\\
  \And
  Yusuke Tanaka\\
  NTT Service Evolution Laboratories\\
}
\date{}

\begin{document}
\maketitle

\begin{abstract}
  We propose a few-shot learning method for spatial regression. Although Gaussian processes (GPs) have been successfully used for spatial regression, they require many observations in the target task to achieve a high predictive performance. Our model is trained using spatial datasets on various attributes in various regions, and predicts values on unseen attributes in unseen regions given a few observed data. With our model, a task representation is inferred from given small data using a neural network. Then, spatial values are predicted by neural networks with a GP framework, in which task-specific properties are controlled by the task representations. The GP framework allows us to analytically obtain predictions that are adapted to small data. By using the adapted predictions in the objective function, we can train our model efficiently and effectively so that the test predictive performance improves when adapted to newly given small data. In our experiments, we demonstrate that the proposed method achieves better predictive performance than existing meta-learning methods using spatial datasets.
\end{abstract}

\section{Introduction}

Thanks to the development of sensors, GPS devices, and satellite systems,
a wide variety of spatial data are being accumulated,
including climate~\cite{stralberg2015projecting,wang2016locally},
traffic~\cite{zheng2014urban,yuan2011t}, economic, and social data~\cite{haining1993spatial,shadbolt2012linked}.
Analyzing such spatial data is critical
in various fields, such as
environmental sciences~\cite{jerrett2005spatial,hession2011spatial},
urban planning~\cite{yuan2012discovering}, socio-economics~\cite{smith2014poverty,rupasingha2007social},
and public security~\cite{bogomolov2014once,wang2016crime}.

Collecting data on some attributes is difficult if the attribute-specific sensing devices are very expensive,
or experts that have extensive domain knowledge are required to observe data.
Also, collecting data in some regions is difficult if they are not readily accessible.
To counter these problems,
many spatial regression methods have been proposed~\cite{gao2006empirical,gao2006estimation,ward2018spatial};
they predict missing attribute values given data observed at some locations in the region.
Although Gaussian processes (GPs)~\cite{williams2006gaussian,banerjee2008gaussian} have been successfully used for spatial regression,
they fail when the data observed in the target region are insufficient.

\begin{figure}[t!]
  \centering
  \includegraphics[width=22em]{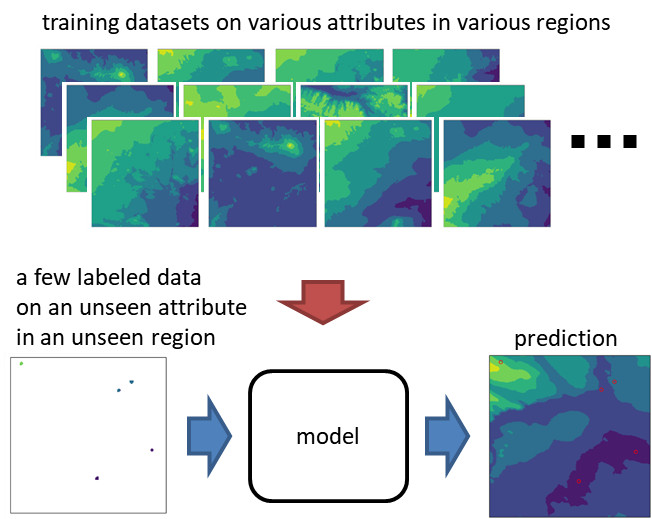}
  \caption{Our framework. In a training phase, our model learns from training datasets containing various attributes from various regions. In a test phase, our model predicts spatial values of a target attribute in a target region given a few observations; the target attribute and region are not present in the training datasets.}
  \label{fig:framework}
\end{figure}

In this paper,
we propose a few-shot learning method for spatial regression.
Our model learns
from spatial datasets on various attributes in various regions,
and predicts values when observed data in the target task is scant,
where both the attribute and region of the target task are different from those in the training datasets.
Figure~\ref{fig:framework} illustrates the framework of the proposed method.
Some attributes in some regions are expected to exhibit similar spatial patterns to the target task.
Our model uses the knowledge learned from such attributes and regions
in the training datasets to realize prediction in the target task.

Our model uses a neural network to embed a few labeled data
into a task representation.
Then, target spatial data are predicted based on a GP
with neural network-based mean and kernel functions
that depend on the inferred task representation.
Using the task representation yields
a task-specific prediction function.
By basing the modeling on GPs,
the prediction function can be rapidly adapted to small labeled data in a closed form
without iterative optimization,
which enables efficient back-propagation through the adaptation.
As the mean and kernel functions employ neural networks,
we can flexibly model spatial patterns in various attributes and regions.
By sharing the neural networks across different tasks in our model, 
we can learn from multiple attributes and regions,
and use the learned knowledge to handle new attributes and regions.
The neural network parameters are estimated
by minimizing the expected prediction performance when a few observed data are given,
which is calculated using training datasets
by an episodic training framework~\cite{ravi2016optimization,santoro2016meta,snell2017prototypical,finn2017model,li2019episodic}.

The main contributions of this paper are as follows:
1) We present a framework of few-shot learning for spatial regression.
2) We propose a GP-based model that uses neural networks to
learn spatial patterns from various attributes and regions.
3) We empirically demonstrate that the proposed method performs well
  in few-shot spatial regression tasks.

\section{Related work}
\label{sec:related}

GPs, or kriging~\cite{cressie1990origins}, have been widely used for spatial
regression~\cite{banerjee2008gaussian,luttinen2009variational,park2011domain,stein2012interpolation,gu2012spatial}.
They achieve high prediction performance at locations that are close to the observed locations.
However, if the target region is large and only a few observed data are given,
performance falls at locations far from the observed locations.
For improving generalization performance,
neural networks have been used
for mean and/or kernel functions of GPs~\cite{wilson2011gaussian,huang2015scalable,calandra2016manifold,wilson2016deep,wilson2016stochastic,iwata2017improving,iwata2019efficient}.
However, these methods require a lot of training data.

Many few-shot learning, or meta-learning, methods have been
proposed~\cite{schmidhuber:1987:srl,bengio1991learning,ravi2016optimization,andrychowicz2016learning,vinyals2016matching,snell2017prototypical,bartunov2018few,finn2017model,li2017meta,kimbayesian,finn2018probabilistic,rusu2018meta,yao2019hierarchically,edwards2016towards,garnelo2018conditional,kim2019attentive,hewitt2018variational,bornschein2017variational,reed2017few,rezende2016one}.
However, they are not intended for spatial regression.
On the other hand, our model is based on GPs, which have been
successfully used for spatial regression.
Some few-shot learning methods based on GPs have been proposed~\cite{harrison2018meta,tossou2019adaptive,fortuin2019deep}.
Adaptive learning for probabilistic connectionist architectures (ALPaCA)~\cite{harrison2018meta}
and adaptive deep kernel learning~\cite{tossou2019adaptive}
incorporate the information in small labeled data in kernel functions using neural networks,
but they assume zero mean functions.
Although meta-learning mean functions~\cite{fortuin2019deep}
use a neural network for the mean function,
the mean function does not change outputs depending on the given small labeled data.
On the other hand, the proposed method uses a neural network-based mean function
that outputs task-specific values by extracting a task representation from the small labeled data.
The effectiveness of our mean function is shown in the ablation study in our experiments.
A summary of related methods is shown in the appendix.

Our model is related to 
conditional neural processes (NPs)~\cite{garnelo2018conditional,garnelo2018neural}
as both use neural networks for task representation inference
and for prediction with inferred task representations.
However, since NP prediction is based on fully parametric models,
they are less flexible in adapting to the given target observations
than GPs, which are nonparametric models.
In contrast, our GP-based model enjoys the benefits of the nonparametric approach,
swift adaptation to the target observations,
even though the mean and kernel functions are modeled parametrically.
Our model is also related to similarity-based meta-learning methods,
such as
matching networks~\cite{vinyals2016matching} and prototypical networks~\cite{snell2017prototypical},
since the kernel function represents similarities between data points.
Although existing similarity-based meta-learning methods were designed for classification tasks,
our model is designed for regression tasks.

The proposed method is also related to model-agnostic meta-learning
(MAML)~\cite{finn2017model}
in the sense that
both methods trains models
so that the expected error on unseen data is minimized
when adapted to a few observed data.
For the adaptation, MAML requires costly back-propagation through iterative gradient descent steps.
On the other hand, the proposed method achieves an efficient adaptation
in a closed form using a GP framework.
Ridge regression differentiable discriminator (R2-D2)~\cite{bertinetto2018meta}
is a neural network-based meta-learning method, where
the last layer is adapted by solving a ridge regression problem in a closed form.
Although R2-D2 and \cite{lee2019meta} adapt with a linear model,
the proposed method adapts with a nonlinear GP model,
which enables us to adapt to complicated patterns more flexibly.

Transfer learning methods, such as
multi-task GPs~\cite{yu2005learning,bonilla2008multi,wei2017source}
and co-kriging~\cite{myers1982matrix,stein1991universal},
have been proposed;
they transfer knowledge derived from source tasks to target tasks.
However, they do not assume a few observations in target tasks.
In addition, since these methods use target data to
learn the relationship between source and target tasks,
they require computationally costly re-training given new tasks that
are not present in the training phase.
On the other hand, the proposed method
can be applied to unseen tasks
by inferring task representations
from a few observations without re-training. 

\section{Proposed method}
\label{sec:proposed}

\subsection{Task}

  \begin{figure}[t!]
  \centering
  \includegraphics[width=25em]{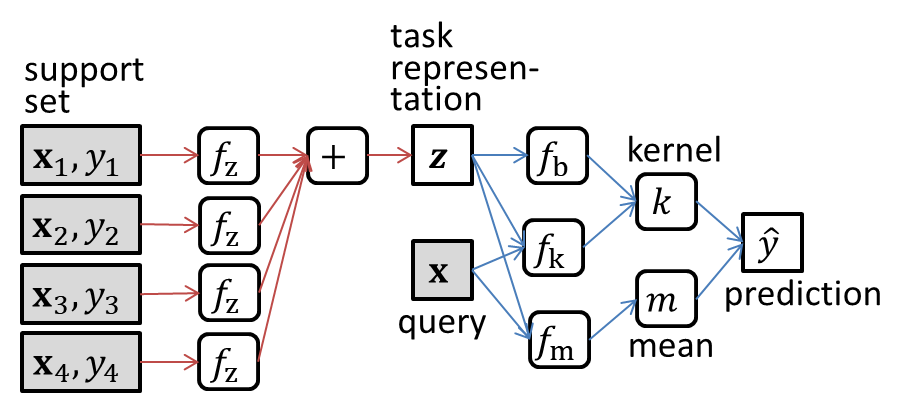}
  \caption{Our model. Each pair of location vector $\vec{x}_{n}$ and attribute value $y_{n}$ in a few labeled data set (support set) is fed to neural network $f_{\mathrm{z}}$. By averaging the outputs of the neural network, we obtain task representation $\vec{z}$. The task representation and query location vector $\vec{x}$ are fed to neural networks $f_{\mathrm{b}}$, $f_{\mathrm{k}}$, and $f_{\mathrm{m}}$ to calculate kernel $k$ and mean $m$. Attribute value $\hat{y}$ of the query location vector is predicted by using the kernel and mean based on a GP. Shaded nodes represent observed data.}
  \label{fig:model}
\end{figure}

In a training phase, we are given spatial datasets for $|\mathcal{R}|$ regions,
$\mathcal{D}=\{\mathcal{D}_{r}\}_{r\in\mathcal{R}}$,
where $\mathcal{R}$ is the set of regions, and
$\mathcal{D}_{r}$ is the dataset for region $r$.
For each region, there are $|\mathcal{C}_{r}|$ attributes,
$\mathcal{D}_{r}=\{\mathcal{D}_{rc}\}_{c\in\mathcal{C}_{r}}$,
where $\mathcal{C}_{r}$ is the set of attributes in region $r$,
and $\mathcal{D}_{rc}$ is the dataset of attribute $c$ in region $r$.
The attribute sets can be different across the regions.
Each dataset consists of
a set of location vectors and attribute values,
$\mathcal{D}_{rc}=\{(\vec{x}_{rcn},y_{rcn})\}_{n=1}^{N_{rc}}$,
where $\vec{x}_{rcn}\in\mathbb{R}^{2}$
is a two-dimensional vector specifying the location of the $n$th point, e.g., longitude and latitude,
and 
$y_{rcn}\in\mathbb{R}$
is the scalar value on attribute $c$ at that location.

In a test phase, we are given
a few labeled observations in a target region,
$\mathcal{D}_{r^{*}c^{*}}=\{(\vec{x}_{r^{*}c^{*}n},y_{r^{*}c^{*}n})\}_{n=1}^{N_{r^{*}c^{*}}}$,
where target region $r^{*}$ is not one of the regions in the training datasets,
$r^{*}\notin\mathcal{R}$,
and target attribute $c^{*}$ is not contained in the training datasets,
$c^{*}\notin\mathcal{C}_{r}$ for all $r\in\mathcal{R}$.
Our task is to predict target attribute value $\hat{y}_{r^{*}c^{*}}$ at location $\vec{x}_{r^{*}c^{*}}$
in the target region.

Location vector $\vec{x}_{rcn}$ represents
the relative position of the point in region $r$.
We used longitudes and latitudes normalized with zero mean for the location vectors in our experiments.
Spatial data sometimes include auxiliary information such as elevation.
In that case, we can include the auxiliary information
in $\vec{x}_{rcn}\in\mathbb{R}^{M+2}$, where $M$ is the number of additional types of auxiliary information.

\subsection{Model}

Let $\mathcal{S}=\{(\vec{x}_{n},y_{n})\}_{n=1}^{N}$
be a few labeled observations, which are called the {\it support set}.
We here present our model for predicting attribute scalar value
$\hat{y}$ at location vector $\vec{x}$,
which is called the {\it query},
given support set $\mathcal{S}$.
This model is used for training as described in Section~\ref{sec:learning}
as well as target spatial regression in a test phase.
Figure~\ref{fig:model} illustrates our model.
Our model infers task representation $\vec{z}$
from support set $\mathcal{S}$ as described in Section~\ref{sec:inference}.
Then, using the inferred task representation $\vec{z}$,
we predict attribute scalar value $\hat{y}$ of location vector $\vec{x}$
by a neural network-based GP as described in Section~\ref{sec:prediction}.
We omit indices for regions and attributes for simplicity in this subsection.

\subsubsection{Inferring task representation}
\label{sec:inference}

First, each pair of the location vector and attribute value, ($\vec{x}_{n},y_{n}$), in the support set
is converted into $K$-dimensional latent vector $\vec{z}_{n}\in\mathbb{R}^{K}$
by a neural network:
$\vec{z}_{n}=f_{\mathrm{z}}([\vec{x}_{n},y_{n}])$,
where $f_{\mathrm{z}}$ is a feed-forward neural network with
the $(M+3)$-dimensional input layer and $K$-dimensional output layer,
and $[\cdot,\cdot]$ represents the concatenation.
Second, the set of latent vectors $\{\vec{z}_{n}\}_{n=1}^{N}$ in the support set
are aggregated to $K$-dimensional latent vector $\vec{z}\in\mathbb{R}^{K}$ by averaging:
$\vec{z}=\frac{1}{N}\sum_{n=1}^{N}\vec{z}_{n}$,
which is a representation of the task extracted from support set $\mathcal{S}$.
We can use other aggregation functions,
such as summation~\cite{zaheer2017deep}, attention-based~\cite{kim2019attentive},
and recurrent neural networks~\cite{vinyals2016matching}.

\subsubsection{Predicting attribute values}
\label{sec:prediction}

Our prediction function assumes
a GP with neural network-based mean and kernel functions
that depend on the inferred task representation $\vec{z}$.
In particular, the mean function is modeled by 
\begin{align}
  m(\vec{x};\vec{z}) = f_{\mathrm{m}}([\vec{x},\vec{z}]),
  \label{eq:mean}  
\end{align}
where
$f_{\mathrm{m}}$ is a feed-forward neural network that outputs a scalar value.
The kernel function is modeled by
\begin{align}
  k(\vec{x},\vec{x}';\vec{z}) &= \exp\left(-\parallel f_{\mathrm{k}}([\vec{x},\vec{z}])-f_{\mathrm{k}}([\vec{x}',\vec{z}])\parallel^{2}
  \right)
  +f_{\mathrm{b}}(\vec{z})\delta(\vec{x},\vec{x}'),
  \label{eq:kernel}
\end{align}
where
$f_{\mathrm{k}}$ is a feed-forward neural network,
$f_{\mathrm{b}}$ is a feed-forward neural network with that outputs
a positive scalar value,
$\delta(\vec{x},\vec{x}')=1$ if $\vec{x}$ and $\vec{x}'$ are identical, and zero otherwise.
The kernel function is positive definite since
it is a Gaussian kernel
and $f_{\mathrm{b}}(\vec{z})$ is positive.
By incorporating task representation $\vec{z}$
in the mean and kernel functions using neural networks,
we can model nonlinear functions that depend on the support set.

In GPs, zero mean functions are often used
since the GPs with zero mean functions can approximate an arbitrary continuous function,
if given enough data~\cite{micchelli2006universal}.
However, GPs with zero mean functions predict zero at areas
far from observed data points~\cite{iwata2017improving},
which is problematic in few-shot learning.
Modeling the mean function by a neural network~(\ref{eq:mean})
allows us to predict values effectively even in areas
far from observed data points in a target region
due to the high generalization performance of neural networks.

Location vector $\vec{x}$ is transformed
by neural network $f_{\mathrm{k}}$ before computing the kernel function by the Gaussian kernel in (\ref{eq:kernel}).
The use of the neural network yields flexible modeling of the
correlation across locations depending on the task representation.
The noise parameter is also modeled by neural network $f_{\mathrm{b}}$,
which enables us to infer the noise level from the support set without re-training.

The predicted value for query $\vec{x}$ is given by
\begin{align}
  \hat{y}(\vec{x},\mathcal{S};\bm{\Phi})=f_{\mathrm{m}}([\vec{x},\vec{z}])+\vec{k}^{\top}\vec{K}^{-1}(\vec{y}-\vec{m}),
  \label{eq:prediction}
\end{align}
where
$\vec{K}$ is the $N\times N$ matrix of the kernel function evaluated between location vectors
in the support set,
$\vec{K}_{nn'}=k(\vec{x}_{n},\vec{x}_{n'})$,
$\vec{k}$ is the $N$-dimensional vector of the kernel function 
between the query and support set, $\vec{k}=(k(\vec{x},\vec{x}_{n}))_{n=1}^{N}$,
$\vec{y}$ is the $N$-dimensional vector of attribute values in the support set,
$\vec{y}=(y_{n})_{n=1}^{N}$,
$\vec{m}$ is the $N$-dimensional vector of the mean function evaluated on locations in the support set,
$\vec{m} = (f_{\mathrm{m}}([\vec{x}_{n},\vec{z}]))_{n=1}^{N}$,
and $\bm{\Phi}$ are the parameters of neural networks $f_{\mathrm{z}}$, $f_{\mathrm{m}}$, $f_{\mathrm{k}}$, and $f_{\mathrm{b}}$.
An advantage of our model is that
the predicted value given the support set is analytically calculated
without iterative optimization,
by which we can minimize the expected prediction error efficiently based on gradient-descent methods.

When noise $f_{\mathrm{b}}(\vec{z})$ is small,
the predicted value approaches the observed values
at locations close to the observed locations.
This property of GPs is beneficial for few-shot regression without re-training.
If a neural network without GPs is used for the prediction function, 
the predicted values might differ from the observations
even at the observed locations
when re-training based on the observations is not conducted.
The first term in (\ref{eq:prediction}) is similar to conditional neural processes,
where a neural network is used for the prediction function.
The second term in (\ref{eq:prediction}) is related to similarity-based
meta-learning methods since the second term uses the similarities between the query and support set
that are calculated by the kernel function.
Therefore, our model can be seen as an extension of
the conditional neural process and similarity-based meta-learning approach,
where both of them are naturally integrated within a GP framework.
When $\vec{x}$ is far from (close to) the observed locations,
the first (second) term becomes dominant due to kernel $\vec{k}$~\cite{iwata2017improving}.
This is reasonable since similarity-based approaches are more reliable
when there are observations nearby.
The variance of the predicted attribute value of the query is given by
$\mathbb{V}[y|\vec{x},\mathcal{S};\bm{\Phi}] = k(\vec{x},\vec{x};\vec{z})-\vec{k}^{\top}\vec{K}^{-1}\vec{k}$.

\subsection{Learning}
\label{sec:learning}

We estimate neural network parameters $\bm{\Phi}$ 
by minimizing the expected prediction error
on a query set given a support set
using an episodic training framework~\cite{ravi2016optimization,santoro2016meta,snell2017prototypical,finn2017model,li2019episodic}.
Although training datasets $\mathcal{D}$ contain many observations,
they should be used in a way that closely simulates the test phase.
Therefore, with the episodic training framework,
support and query sets are generated
by a random subset of training datasets $\mathcal{D}$
for each training iteration.
In particular, we use the following objective function:
\begin{align}
  \hat{\bm{\Phi}}=
  \arg \min_{\bm{\Phi}} \mathbb{E}_{r\sim\mathcal{R}}[\mathbb{E}_{c\sim\mathcal{C}_{r}}[
      \mathbb{E}_{(\mathcal{S},\mathcal{Q})\sim\mathcal{D}_{rc}}[
      L(\mathcal{S},\mathcal{Q};\bm{\Phi})]],
\end{align}
where $\mathbb{E}$ represents an expectation,
\begin{align}
  L(\mathcal{S},\mathcal{Q};\bm{\Phi})
  =  \frac{1}{N_{\mathrm{Q}}}
  \sum_{(\vec{x},y)\in\mathcal{Q}}
  \parallel \hat{y}(\vec{x},\mathcal{S};\bm{\Phi})-y \parallel^{2},
  \label{eq:E}
\end{align}
is the mean squared error on query set $\mathcal{Q}$ given support set $\mathcal{S}$,
and $N_{\mathrm{Q}}$ is the number of instances in the query set.
Usually, GPs are trained by maximizing the marginal likelihood
of training data (support set),
where test data (query set) are not used. 
On the other hand,
the proposed method minimizes the prediction error
on a query set when a support set is observed,
by which we can simulate a test phase and
learn a model that improves the prediction performance on target tasks.
When we want to predict the density, we can use the following negative predictive
log-likelihood
\begin{align}
  L(\mathcal{S},\mathcal{Q};\bm{\Phi})
  =  -\frac{1}{N_{\mathrm{Q}}}
  \sum_{(\vec{x},y)\in\mathcal{Q}}
  \mathcal{N}(y|\hat{y}(\vec{x},\mathcal{S};\bm{\Phi}),\mathbb{V}[y|\vec{x},\mathcal{S};\bm{\Phi}]),
  \label{eq:E2}
\end{align}
instead of the mean squared error (\ref{eq:E}),
where $\mathcal{N}(y|\mu,\sigma^{2})$ is the Gaussian distribution with mean $\mu$ and variance $\sigma^{2}$.
This is related to training GPs with the log pseudo-likelihood~\cite{williams2006gaussian},
where the leave-one-out predictive log-likelihood is used as the objective function.

The training procedure of our model is shown in
Algorithm~\ref{alg}.
The computational complexity for evaluation loss (\ref{eq:E})
is $O(N_{\mathrm{Q}}N_{\mathrm{S}}^{3})$, where $N_{\mathrm{S}}$ is
the number of instances in the support set
since we need the inverse of the kernel matrix whose size is $N_{\mathrm{S}}\times N_{\mathrm{S}}$.
In few-shot learning, the number of target observed data is very small,
and so a very small support size $N_{\mathrm{S}}$ is used in training.
Therefore, our model can be optimized efficiently
with the episodic training framework.
This is in contrast to the high computational complexity of training for standard GP regression, 
which is cubic in the number of training instances.

    \begin{algorithm}[t]
      \caption{Training procedure of our model.}
      \label{alg}
      \begin{algorithmic}[1]
        \renewcommand{\algorithmicrequire}{\textbf{Input:}}
        \renewcommand{\algorithmicensure}{\textbf{Output:}}
        \REQUIRE{Spatial datasets $\mathcal{D}$,
          support set size $N_{\mathrm{S}}$, query set size $N_{\mathrm{Q}}$}
        \ENSURE{Trained neural network parameters $\bm{\Phi}$}
        \WHILE{not done}
        \STATE Randomly sample region $r$ from $\mathcal{R}$
        \STATE Randomly sample attribute $c$ from $\mathcal{C}_{r}$
        \STATE Randomly sample support set $\mathcal{S}$ from $\mathcal{D}_{rc}$
        \STATE Randomly sample query set $\mathcal{Q}$ from $\mathcal{D}_{rc}\setminus\mathcal{S}$    
        \STATE Calculate loss by Eq.~(\ref{eq:E}) or (\ref{eq:E2}), and its gradients
        \STATE Update model parameters $\bm{\Phi}$ using the loss and its gradients
        \ENDWHILE
      \end{algorithmic}
    \end{algorithm}

\section{Experiments}
\label{sec:experiments}

\subsection{Data}

We evaluated the proposed method using the following three spatial datasets:
NAE, NA, and JA.
NAE and NA were
the climate data in North American, which were obtained from \url{https://sites.ualberta.ca/~ahamann/data/climatena.html}.
As the location vector,
NA used longitude and latitude.
With NAE, elevation in meters above sea level was additionally used in the location vector.
For attributes, we used 26 bio-climate values,
such as the annual heat-moisture index and climatic moisture deficit.
We generated 1829 non-overlapping regions covering North America,
where the size of each region was 100km $\times$ 100km,
and attribute values were observed at 1km $\times$ 1km grid squares
in each region.
300 training, 18 validation, and 549 target regions
were randomly selected without replacement from the 1829 regions.
Also, 26 attributes were randomly splitted into
20 training, five validation, and one target attributes.
JA was the climate data in Japan, which was obtained from
\url{http://nlftp.mlit.go.jp/ksj/gml/datalist/KsjTmplt-G02.html}.
We used seven climate attributes, such as
precipitation and maximum temperature.
The data contained 273 regrions,
where attribute values were observed at 1km $\times$ 1km grid square,
and there were at most 6,400 locations in a region.
The 273 regions were randomly splitted into 
200 training, 28 validation, 45 target regions.
Also, seven attributes were randomly splitted into
four training, two validation, and one target attributes.
The details of the datasets were described in the appendix.
For all data, in each target region,
values on a target attribute at five locations were observed,
and values at the other locations were used for evaluation.

\subsection{Results}

We compared the proposed method
with the conditional neural process (NP),
Gaussian process regression (GPR),
neural network (NN),
fine-tuning with NN (FT), and
model-agnostic meta-learning with NN (MAML).
The detailed settings of the proposed method and comparing methods
are described in the appendix.

\begin{table}[t!]
  \centering
  \caption{(a) Test mean squared errors and (b) test log likelihoods averaged over ten experiments. Values in bold typeface are not statistically significantly different at the 5\% level from the best performing method in each row according to a paired t-test.}
  \label{tab:err}
  \begin{tabular}{cc}
    \begin{minipage}{0.5\textwidth}
      \centering
        (a) Test mean squared errors\\
        {\tabcolsep=0.25em
          \begin{tabular}{lrrrrrr}
        \hline
        Data & Ours & NP & GPR & NN & FT & {\small MAML}\\
        \hline
        NAE & {\bf 0.316} & 0.348 & 0.476 & 0.963 & 0.497 & 0.710 \\
        NA & {\bf 0.552} & 0.593 & 0.701 & 0.986 & 0.746 & 0.824 \\
        JA & {\bf 0.653} & 0.756 & 0.703 & 1.016 & 0.871 & 0.873 \\ 
        \hline
        \end{tabular}}
      \end{minipage}
    \begin{minipage}{0.5\textwidth}
      \centering      
      (b) Test log likelihoods\\
        {\tabcolsep=0.25em      
      \begin{tabular}{rrrrrr}
        \hline
        Ours & NP & GPR & NN & FT & {\small MAML}\\
        \hline
        {\bf -0.987} & -1.025 & -1.067 & -1.281 & -1.271 & -1.256 \\ 
        {\bf -1.166} & -1.190 & -1.230 & -1.322 & -1.311 & -1.306 \\
        {\bf -1.227} & -1.307 & {\bf -1.220} & -1.333 & -1.598 & -1.314 \\
        \hline
      \end{tabular}}
    \end{minipage}
  \end{tabular}
\end{table}

The test mean squared errors (a) and
test log likelihoods (b) in the target tasks
averaged over ten experiments are shown in Table~\ref{tab:err}.
Here, for the test mean squared error evaluations,
all methods were trained with the mean squared error objective function in Eq.~(\ref{eq:E}),
and for the test log likelihood evaluations,
all methods were trained with the negative log likelihood objective function in Eq.~(\ref{eq:E2}).
The proposed method achieved the best performance in all cases except for
the test likelihood with JA data.
NP was worse than the proposed method
because its predictions were poor
when task representations were not properly inferred.
On the other hand, 
the proposed method performed well with any tasks in at least areas close to the observations,
as its GP framework offers a smooth nonlinear function that passes over the observations.
GPR was worse than the proposed method
since GPR only shares kernel parameters across different tasks.
In contrast,
the proposed method shares neural networks across different tasks,
which enables us to
learn flexible spatial patterns in various attributes and regions
and use them for target tasks.
NN suffered the worst performance since it cannot use the target data.
Fine-tuning (FT) decreased the error,
but it remained worse than that of the proposed method.
This is because
FT consisted of two separate steps: pretrain and fine-tuning,
and did not learn how to transfer knowledge.
In contrast, the proposed method trained the neural network
in a single step so that test performance
is maximized when the support set is given in the episodic training framework.
MAML performance was low
since it was difficulty in learning the parameters
that fine-tuned well with just a small number of epochs
with various regions and attributes,
where target function shapes vary drastically.
Note that due to the high computational complexity of MAML,
where it demands that the gradients of many gradient-descent steps be calculated,
MAML makes it infeasible to use a large number of fine-tuning epochs.
On the other hand,
with the proposed method,
since predicted values given the support set
are calculated analytically based on a GP,
the neural networks are optimized efficiently in terms of fitting the support set,
and therefore the trained model
attained high prediction performance for various attributes and regions.
Since the number of training attributes was small with JA data,
and training data were insufficient to train neural networks,
the test likelihood of the proposed method was not significantly different from that of GPR.

\begin{figure*}[t!]
  \centering
  {\tabcolsep=0em
  \begin{tabular}{ccc}
  \includegraphics[width=15em]{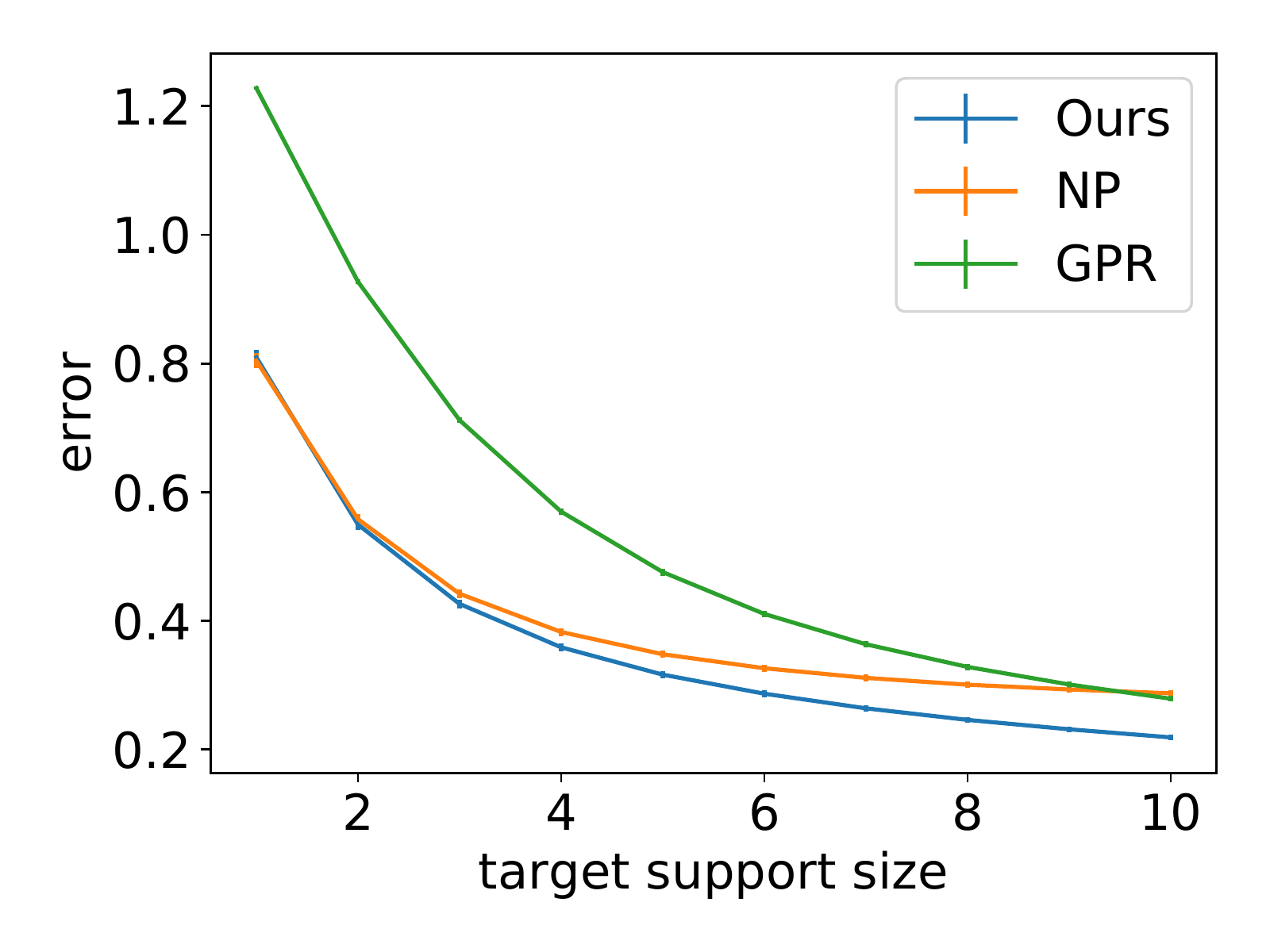}&
  \includegraphics[width=15em]{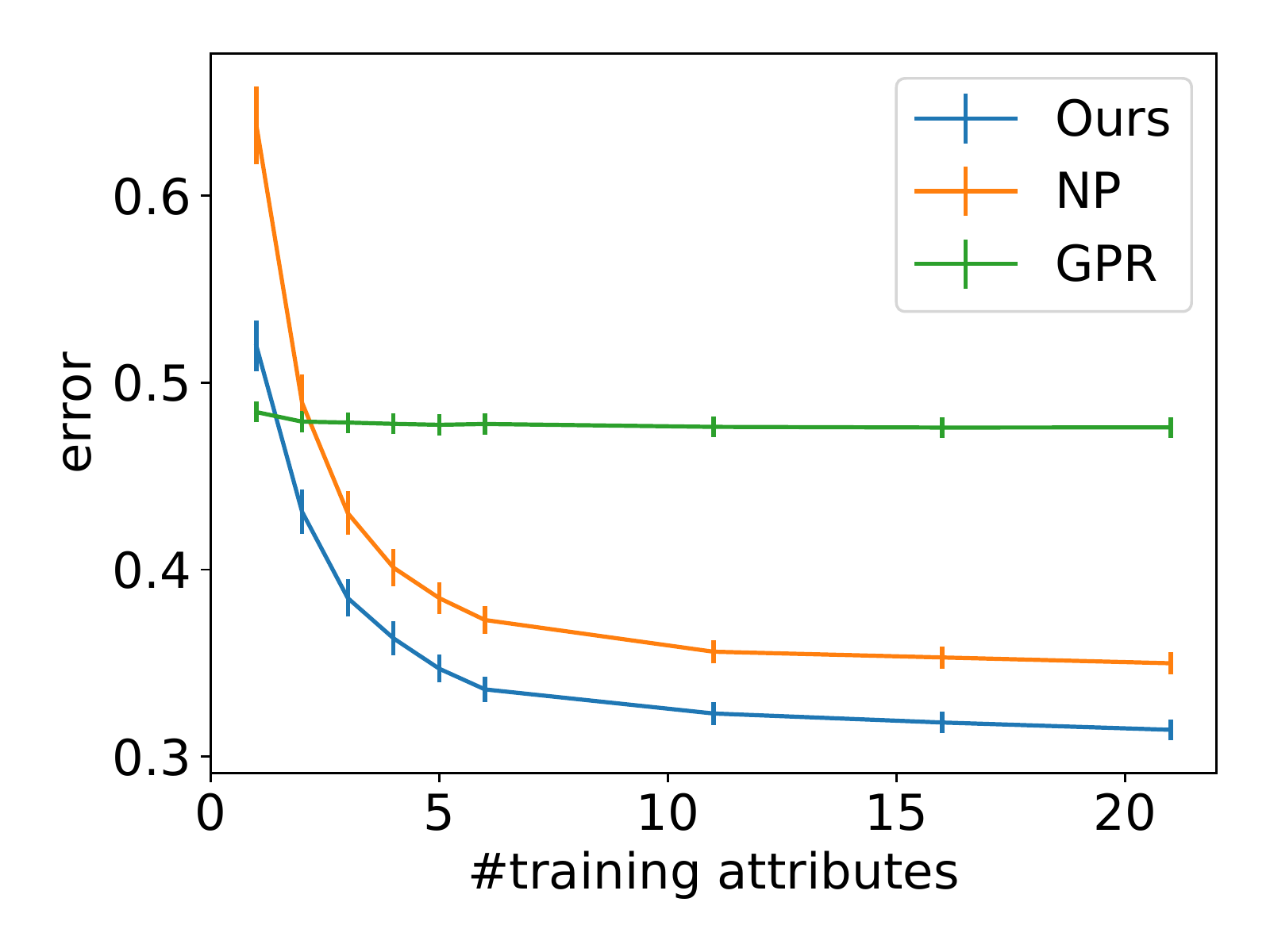}&
  \includegraphics[width=15em]{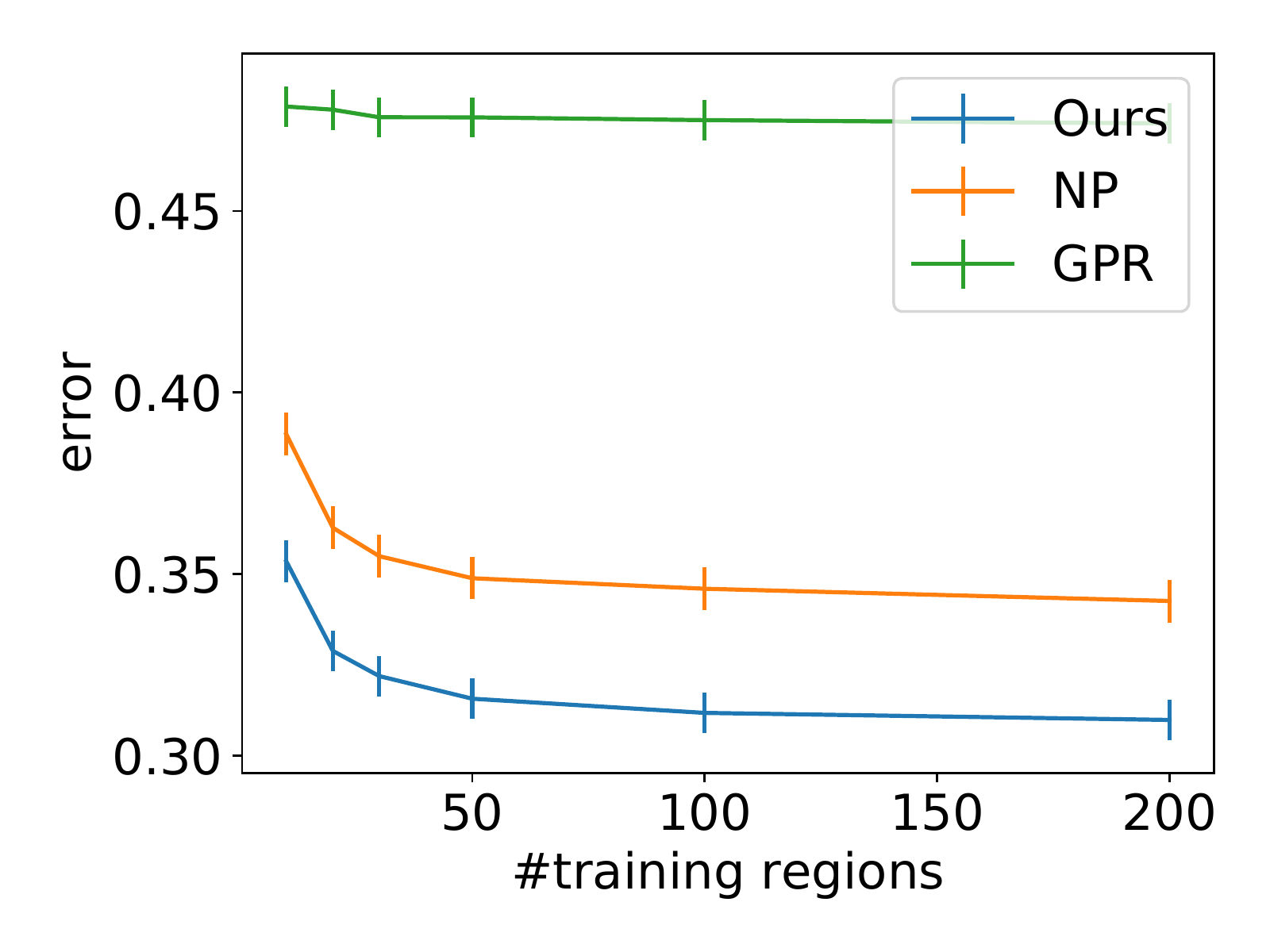}
  \end{tabular}}
  \caption{Average test mean squared errors with (a) different target support sizes, (b) different numbers of training attributes, and (c) different numbers of training regions on NAE data. The bar shows the standard error.}
  \label{fig:mse}
\end{figure*}

Figure~\ref{fig:mse}(a)
shows the average test mean squared errors with different numbers of target support sizes
with the proposed method, NP, and GPR.
We omitted the results with NN, FT, and MAML since
their performance was low as shown in Table~\ref{tab:err}.
All methods yielded decreased error
as the target support size increased.
The proposed method achieved low errors with different target support sizes
since it uses neural networks to learn the relationship
between support and query sets using the training datasets.
NP achieved low error rates when the target support size was small.
However, NP had higher error rates than GPR when the size was ten.
Since NP used a fixed trained neural network to incorporate the support set information,
it was difficult to adapt prediction functions to a large support set.
In contrast, since the proposed method and GPR
can adapt them easily to support sets by calculating the posterior in a closed form,
their errors were effectively decreased as the target support size increased.

Figure~\ref{fig:mse}(b) shows
the average test mean squared errors with different numbers of training attributes.
The errors with the proposed method and NP decreased
as the training attribute numbers increased.
This is reasonable since
the possibility that tasks similar to target tasks
are included in the training datasets
increases as the number of training attributes increases.
Since GPR shared only kernel parameters
across different tasks,
its performance was not improved even when many attributes were used.
Figure~\ref{fig:mse}(c) shows
the average test mean squared errors with different numbers of training regions.
The errors with the proposed method and NP decreased
as the training regions increased.
The computational time of the proposed and comparing methods was described in the appendix.

\begin{figure*}[t!]
  \centering
      {\tabcolsep=0em\begin{tabular}{ccccccc}
       \multicolumn{5}{c}{True attribute values}\\
    \includegraphics[width=8em]{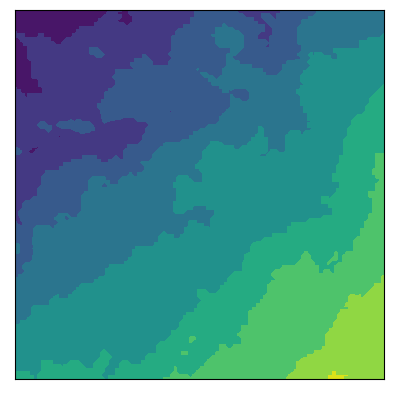}&
    \includegraphics[width=8em]{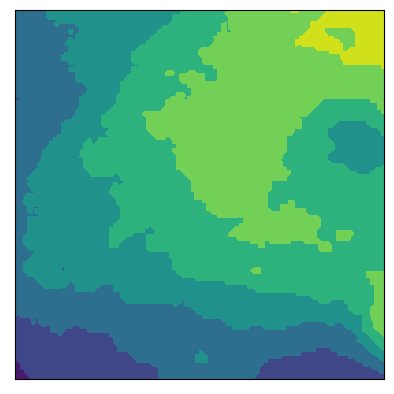}&
    \includegraphics[width=8em]{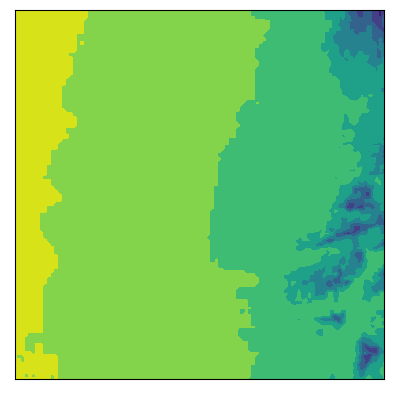}&
    \includegraphics[width=8em]{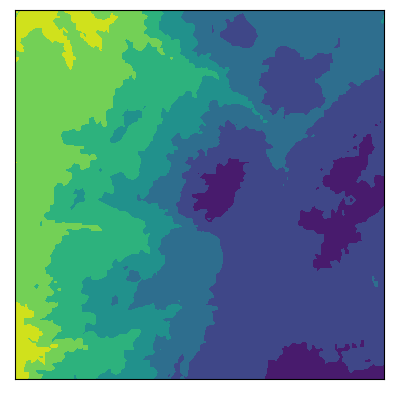}&
    \includegraphics[width=8em]{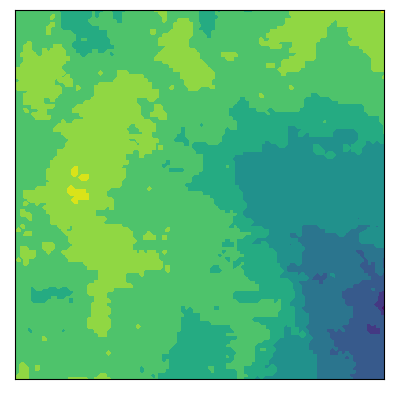}\\
       \multicolumn{5}{c}{Ours}\\
    \includegraphics[width=8em]{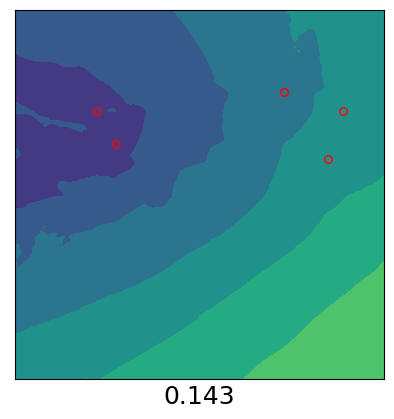}&
    \includegraphics[width=8em]{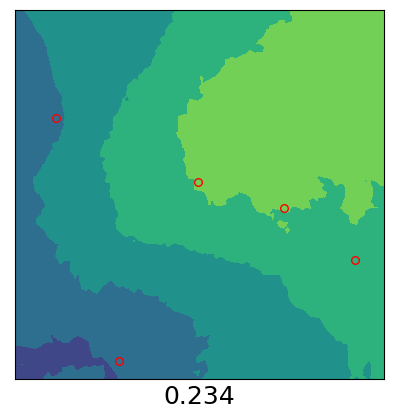}&
    \includegraphics[width=8em]{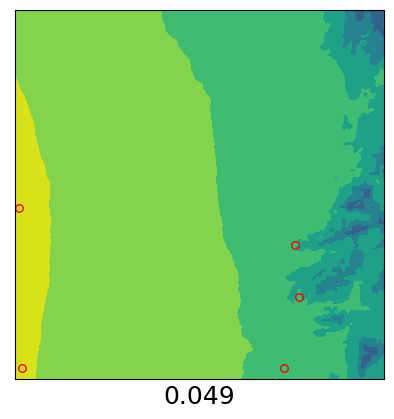}&
    \includegraphics[width=8em]{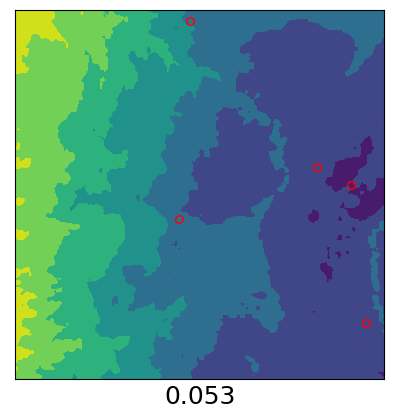}&
    \includegraphics[width=8em]{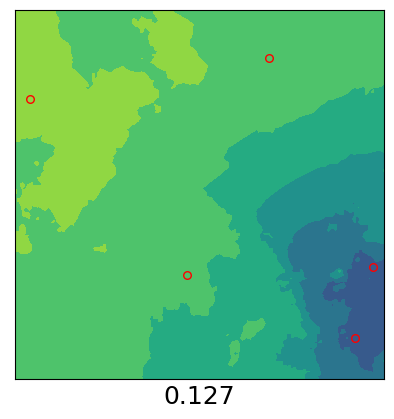}\\
       \multicolumn{5}{c}{NP}\\
    \includegraphics[width=8em]{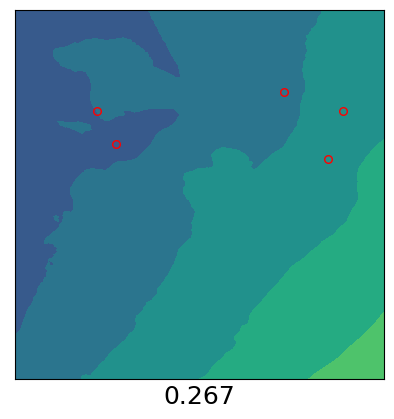}&
    \includegraphics[width=8em]{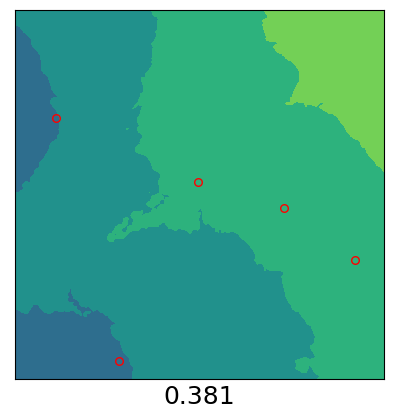}&
    \includegraphics[width=8em]{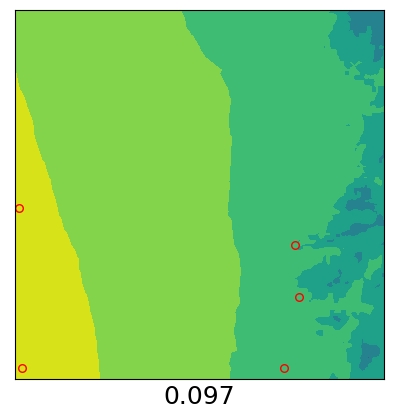}&
    \includegraphics[width=8em]{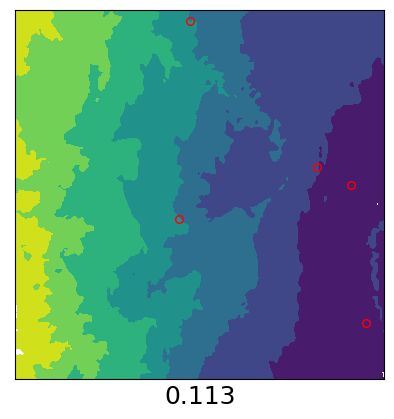}&
    \includegraphics[width=8em]{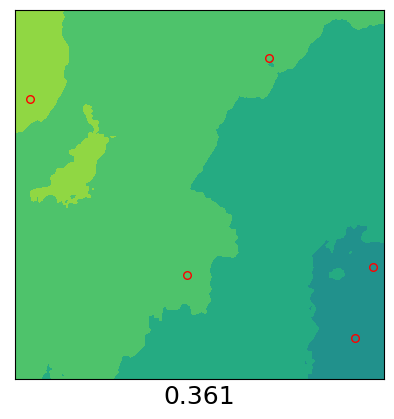}\\
       \multicolumn{5}{c}{GPR}\\
    \includegraphics[width=8em]{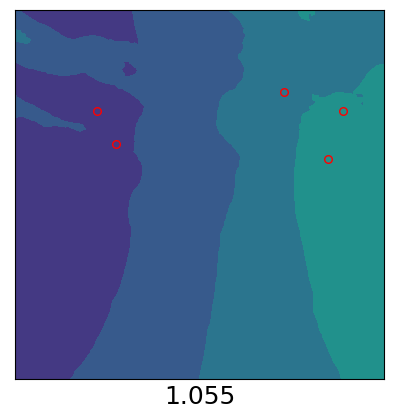}&
    \includegraphics[width=8em]{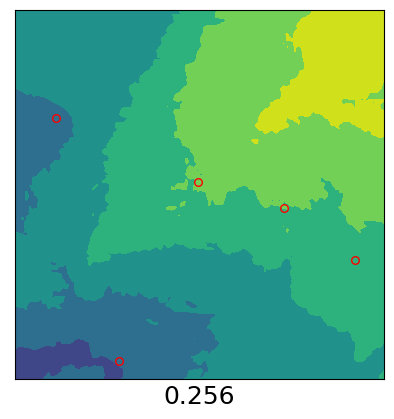}&
    \includegraphics[width=8em]{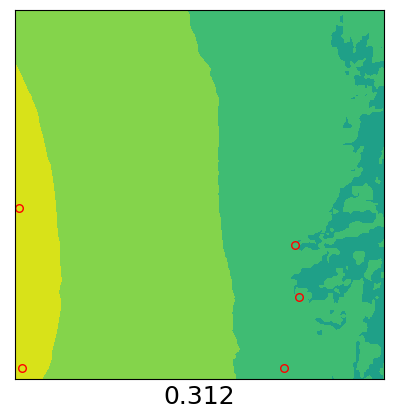}&
    \includegraphics[width=8em]{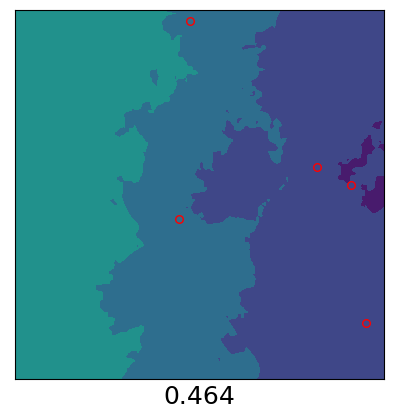}&
    \includegraphics[width=8em]{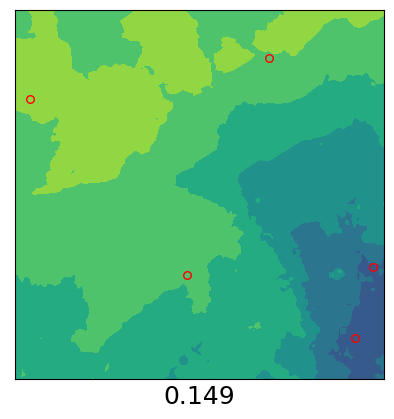}\\
    (a) PAS &
    (b) PPTsm & (c) EMT & (d) DD18 & (e) bFFP \\
    \\
  \end{tabular}}
      \caption{Predictions for five attributes and regions of
        target tasks yielded by the proposed method, NP, and GPR.
        The top row shows the true attribute values.
        Red circles indicate observed locations. Values below each plot shows the mean squared error.}
  \label{fig:viz}
\end{figure*}

Figure~\ref{fig:viz} visualizes the predictions
for five attributes and regions
of target tasks
with the proposed method, NP, and GPR.
The proposed method attained appropriate predictions in various attributes and regions.
NP did not necessarily
output predicted values that were similar to the observations.
For example, in Figure~\ref{fig:viz}(a,NP),
the predicted values of NP at two left observed locations
differed from the true value.
On the other hand, the proposed method and GPR
predicted values similar to the observation at the locations.
Since GPR could not extract the rich knowledge present in the training datasets,
it sometimes failed to predict values.
For example, in Figure~\ref{fig:viz}(a,GPR),
the predicted values differed from the true values in the lower area.
In contrast,
the proposed method and NP predicted values at the area well
using neural networks.
The proposed method improved prediction performance by adopting
both advantages of GPs and neural networks.

\begin{table}[t!]
  \centering
  \caption{Ablation study. Test mean squared errors and test log likelihoods on target tasks with NAE data. ErrObj is the proposed method with the mean squared error objective function in Eq.~(\ref{eq:E}). LikeObj is that with the log likelihood objective function in Eq.~(\ref{eq:E2}), MarLikeObj is that with the marginal likelihood on the support set, NoSptMean is that with neural netwok-based mean function without the support information, $m(\vec{x})=f_{\mathrm{m}}(\vec{x})$, and
    ZeroMean is that with zero mean function.}
  \label{tab:ablation}
  \begin{tabular}{lrrrrr}
    \hline
    Evaluation measurement & ErrObj & LikeObj & MarLikeObj & NoSptMean & ZeroMean \\
    \hline
    Test mean squared error & {\bf 0.316} & 0.329 & 0.476 & 0.386 & 0.394\\
    Test log likelihood & -1.025 & {\bf -0.987} & -1.479 & -1.086 & -1.088\\
    \hline
\end{tabular}
\end{table}

Table~\ref{tab:ablation} shows the results of the ablation study of the proposed method.
In terms of the test mean squared error,
the proposed method with the mean squared error objective function (ErrObj) was better than
that with the likelihood objective function (LikeObj).
In terms of the test log likelihood, LikeObj was better than ErrObj.
These results imply that the objective function should be selected properly depending on the applications.
The proposed method with the marginal likelihood objective function on the support set (MarLikeObj)
was worse than ErrObj and LikeObj.
This result demonstrates the effectiveness to use the test performance for the objective function
for few-shot learning although standard GPs are usually trained with the marginal likelihood of training data.
The proposed with mean function without the support information (NoSptMean)
and that with zero mean function (ZeroMean)
performed worse than the proposed method.
This result indicates the importance to use non-zero mean functions that incorporate the support information,
and the advantage of the proposed method over existing GP-based meta-leaning methods
those that use zero mean functions~\cite{harrison2018meta,tossou2019adaptive}
and those that do not use the support information~\cite{harrison2018meta}.

\section{Conclusion}
\label{sec:conclusion}

We proposed a few-shot learning method for spatial regression.
The proposed method can predict
attribute values given a few observations,
even if the target attribute and region are not included in the training
datasets.
The proposed method
uses a neural network to infer a task representation from a few observed data.
Then, it uses the inferred task representation to calculate the predicted values
by a neural network-based Gaussian process framework.
Experiments on climate spatial data
showed that the proposed method
achieved better prediction performance than existing methods.
For future work, we want to apply our framework
to other types of tasks, such as spatio-temporal regression,
regression for non-spatial data, and classification.

\bibliographystyle{abbrv}
\bibliography{neurips_2020spatial}

\newpage
\appendix

\section{Related work}

Table~\ref{tab:comparison} summarizes the comparison of related methods.

\begin{table}[h]
  \centering
  \caption{Comparison of related methods: the proposed method (Ours), Adaptive learning for probabilistic connectionist architectures (ALPaCA)~\cite{harrison2018meta}, adaptive deep kernel learning (ADKL)~\cite{tossou2019adaptive}, deep kernel learning (DKL)~\cite{wilson2016deep}, meta-learning mean functions (MLMF)~\cite{fortuin2019deep}, Gaussian processes (GP), and conditional neural processes (NP)~\cite{garnelo2018conditional}. The objective function column indicates that the method uses the expected predictive performance objective function (Predictive), or the marginal likelihood objective function (Marginal). The mean function column indicates that the method uses zero mean function (Zero), neural network-based mean function without use support sets (Neural), neural network-based mean function with support sets (NeuralSpt). The kernel function columns indicates that the method does not use kernel functions (No), uses neural network-based kernel function without support sets (Neural), or uses neural network-based kernel function with support sets (NeuralSpt).}
  \label{tab:comparison}
  \begin{tabular}{lrrrr}
    \hline
  & Objective function & Mean function & Kernel function \\
    \hline
    Ours & Predictive & NeuralSpt & NeuralSpt \\
  ALPaCA & Predictive & Zero & NeuralSpt \\
  ADKL & Predictive & Zero & NeuralSpt \\
  DKL & Marginal & Zero & Neural \\
  MLMF & Marginal & Neural & NoNeural \\
  GP & Marginal & Zero & NoNeural \\
  NP & Predictive & NeuralSpt & No \\
  \hline
  \end{tabular}
\end{table}

\section{Proposed method setting}

As the neural networks in our model, $f_{\mathrm{z}}$,
$f_{\mathrm{b}}$, $f_{\mathrm{k}}$, and $f_{\mathrm{m}}$,
we used three-layered feed-forward neural networks with
256 hidden units.
The dimensionality of the output layer
with $f_{\mathrm{z}}$ and $f_{\mathrm{k}}$ was 256,
and that with $f_{\mathrm{b}}$ and $f_{\mathrm{m}}$ was one.
We used rectified linear unit, $\mathrm{ReLU}(x)=\max(0,x)$, for activation.
Optimization was performed using Adam~\cite{kingma2014adam} with learning rate $10^{-3}$ and
dropout rate $0.1$.
The maximum number of training epochs was 5,000,
and the validation datasets were used for early stopping.
The support set size was $N_{\mathrm{S}}=5$,
and query set size was $N_{\mathrm{Q}}=64$.

\section{Comparison methods}

We compared the proposed method
with conditional neural process (NP),
Gaussian process regression (GPR),
neural network (NN),
fine-tuning with NN (FT), and
model-agnostic meta-learning with NN (MAML).

With NP, a task representation was
inferred from the support set
using a neural network as in the proposed method,
and then the attribute values of queries were predicted
using another neural network.
We used the same neural network architecture with the proposed method
for inferring task representations $f_{\mathrm{z}}$.
The architecture of the neural network for prediction was the same
as that with
$f_{\mathrm{m}}$ in the proposed method, which was used as the mean function.

GPR predicted the attribute values by
a GP regression with a Gaussian kernel given the support set.
The kernel parameters, which were the signal variance,
length scale, and noise variance, were estimated from the training datasets
by minimizing the expected prediction error
using the episodic training framework.

NN used a three-layered feed-forward neural network with 256 hidden units,
and the ReLU activation was used. The input of the NN was a location vector,
and its output was the predicted value of the attribute.
NN parameters, which were shared across all tasks,
were estimated using the training datasets.
The NN did not use labeled data in target tasks.

FT fine-tuned the parameters of the trained NN with labeled data
for each target task.
For fine-tuning, we used Adam with learning rate $10^{-3}$.
The number of epochs for fine-tuning was 100,
which was selected from $\{10,100\}$ based on the target performance.

MAML used the same neural network as NN.
The parameters were trained so that
the prediction performance was improved
when fine-tuned with a support set.
The number of fine-tuning epochs was five.
MAML was implemented with Higher, which
is a library for higher-order optimization~\cite{grefenstette2019generalized}.

NP, GPR, NN, and MAML used the episodic training framework
in the same way as the proposed method.
All the methods were optimized with Adam with learning rate $10^{-3}$,
and implemented with PyTorch~\cite{paszke2017automatic}.

\section{Data}

With NAE and NA data,
we used the following 26 bio-climate values as attributes:
annual heat-moisture index (AHM),
climatic moisture deficit (CMD),
degree-days above 18$^\circ$C (DD18),
degree-days above 5$^\circ$C (DD5),
degree-days below 0$^\circ$C (DD-0),
degree-days below 18$^\circ$C (DD-18),
extreme minimum temperature over 30 years (EMT),
extreme maximum temperature over 30 years (EXT),
monthly reference evaporation (Eref),
length of the frost-free period (FFP),
mean annual precipitation (MAP),
mean annual temperature (MAT),
mean temperature of the coldest month (MCMT),
mean summer (May to Sep) precipitation (MSP),
mean temperature of the warmest month (MWMT),
the number of frost-free days (NFFD), 
precipitation as snow (PAS),  
summer (Jun to Aug) precipitation (PPT-sm).  
winter (Dec to Feb) precipitation (PPT-wt),
monthly average relative humidity (RH),
summer heat-moisture index (SHM),
difference between MCMT and MWMT, as a measure of continentality (TD),
summer (Jun to Aug) mean temperature (Tave-sm), 
winter (Dec to Feb) mean temperature (Tave-wt), 
the beginning of the frost-free period (bFFP), 
and the end of the frost-free period (eFFP). 
With JA data,
we used the following seven climate values as attributes:
precipitation, maximum temperature, minimum temperature,
average temperature, maximum snow depth, sunshine duration,
and total solar radiation.
The location vectors and attributes were
normalized with zero mean and one standard deviation
for each region and for each attribute.

\section{Computational time}

Table~\ref{tab:time} shows
the average computation time in seconds for learning from the training datasets and the time for predicting test attribute values for each region
on computers with 2.60GHz CPUs.
Although the proposed method had slightly longer training time than
NP, GPR, or NN since it uses both the neural networks and GP,
it was faster than MAML.
All methods had short test times since
the number of observed locations was small.
The proposed method had shorter test time than FT because
the proposed method calculated predictions
analytically given the target data, while
the FT required multiple updates for optimization
given the target data.

\begin{table}[t!]
  \centering
  \caption{Average computational time in seconds for learning from the training datasets and the time for predicting test attribute values for each region.}
  \label{tab:time}
  {\tabcolsep=0.45em
  \begin{tabular}{lrrrrrr}
    \hline
    & Ours & NP & GPR & NN & FT & {\small MAML}\\
    \hline
    Train & 2253.4 & 1031.9 & 637.6 & 756.5 & 756.5 & 7618.8 \\
    Test & 0.142 & 0.050 & 0.005 & 0.026 & 0.161 & 0.067\\
    \hline
  \end{tabular}}
\end{table}

\end{document}